# Posture Prediction for Healthy Sitting using a Smart Chair

Tariku Adane Gelaw[1, 2 [0000-0002-1493-9233]], Misgina Tsighe Hagos [3, 4 [0000-0002-9318-9417]]

[1] University of Trento, Department of Information Engineering and Computer Science, Italy
[2] Fondazione Bruno Kessler, Italy
tarikuadane.gelaw@gmail.com
[3] Science Foundation Ireland Centre for Research Training in Machine Learning at University College Dublin, Ireland
[4] School of Computer Science, University College Dublin, Ireland
misgina.hagos@ucdconnect.ie

**Abstract.** Poor sitting habits have been identified as a risk factor to musculo-skeletal disorders and lower back pain especially on the elderly, disabled people, and office workers. In the current computerized world, even while involved in leisure or work activity, people tend to spend most of their days sitting at computer desks. This can result in spinal pain and related problems. Therefore, a means to remind people about their sitting habits and provide recommendations to counterbalance, such as physical exercise, is important. Posture recognition for seated postures have not received enough attention as most works focus on standing postures. Wearable sensors, pressure or force sensors, videos and images were used for posture recognition in the literature. The aim of this study is to build Machine Learning models for classifying sitting posture of a person by analyzing data collected from a chair platted with two 32 by 32 pressure sensors at its seat and backrest. Models were built using five algorithms: Random Forest (RF), Gaussian Naïve Bayes, Logistic Regression, Support Vector Machine and Deep Neural Network (DNN). All the models are evaluated using KFold cross validation technique. This paper presents experiments conducted using the two separate datasets, controlled and realistic, and discusses results achieved at classifying six sitting postures. Average classification accuracies of 98% and 97% were achieved on the controlled and realistic datasets, respectively.

**Keywords:** Sitting posture, Smart chair, Pressure sensor, Deep Neural Networks, Prediction.

## 1 Introduction

Many people spend a large portion of their daily time, sitting in an office chair, lounge chair, car seat or on wheelchairs. Due to this reason, seat comfort has gained particular attention for nursing homes, military, workplace, and assistive technology applications. Sore muscles, heavy legs, uneven pressure, stiffness, restlessness, fatigue, and pain was considered as symptoms which are caused due to seating discomfort in the office environment [1, 2, 3].

Everyone needs to be more active but at the same time they also wanted to spend less time sitting down. A person in the developed world who primarily works using a computer could reasonably sit for up to 15 hours a day [4]. Other studies also showed that some older adults (aged 65 and over) spend 10 hours or more each day sitting or lying down, making them the most sedentary population group. All of this sitting comes with significant health costs, both from inactivity and from poor posture.

Long periods of sitting have been linked to obesity, cardiovascular disease and premature mortality. Although there is also evidence that these adverse effects can be mitigated by short standing breaks [5, 6]. Poor sitting posture has been identified as a risk factor for musculoskeletal disorders [7], and particularly for lower back pain. Musculoskeletal disorders can cause chronic pain in the limbs, neck and back. In general, this shows us that postural imbalance has a great impact on the health of individuals by causing different diseases. Specially, senior citizens (older adults) and people with disabilities are impacted with this. So, finding a solution to improve postural imbalance of an individual is the key for achieving a healthy and active life. As a result, the aim of this study is also to focus on finding ways to prevent or treat the postural distortions that has a greater impact on the individuals' health and daily activities.

This paper is structured into five sections. Section 2 presents related works found in the literature and the research gap that we worked on in this paper. Our proposed methodology is discussed in Section 3. Experiment results are presented in Section 4. Finally, conclusion of our work and recommendations for future work are presented in Section 5.

## 2 Related Work

A study by [8] used 19 pressure sensors to identify sitting postures. This study used an approximation algorithm for a near-optimal sensor placement. The researchers used a dataset which contains pressure data for ten postures, collected from 52 participants. In this study the researchers achieved an accuracy of 82%. Another research by Kazuhiro et al. [9] used a pressure sensor seat on a chair for identifying sitting postures. In their experiments, [9] classified nine postures, including leaning forward / backward / right / left and legs crossed. In this study, they obtained a classification accuracy of 98.9% when the sitting person was known and 93.9% when the person was not known. A study by [10] also designed a personalized smart chair system to recognize sitting behaviors. The system can receive surface pressure data from the designed sensor and provide feedback for guiding the user towards proper sitting postures. They used a liquid state machine and a Logistic Regression (LR) classifier to construct a spiking neural network for classifying 15 sitting postures. The experimental results consisting of 15 sitting postures from 19 participants show that a prediction precision of 88.52%. Yong et al. developed a system for classifying sitting postures for children using CNN, Naïve Bayes, Decision Tree (DT), Multinomial

Logistic Regression (MLR), Neural Network (NN), and Support Vector Machine (SVM) machine learning algorithms [11]. Ten children participated in this research and achieved an accuracy of 95.3% using CNN. Another researcher [12] proposed a system that uses a specialized Arduino-based chair to predict and analyze the sitting posture of the user and provides appropriate videos to help them correct their posture by analyzing the user statistics on their overall posture data. They used deep CNNs and LBCNet (Lower-Balanced Check Network).

A research conducted by Griffiths et al. [13], in a laboratory study with 18 participants, evaluated a range of common sitting positions to determine when heart rate and respiratory rate detection was possible and evaluate the accuracy of the detected rate. Griffiths et al. employ conductive fabric on the chair's armrests to sense heart rate and pressure sensors on the back of the chair for sensing respiratory rate. Arnrich et al. used detected chair information for understanding stress level of individuals during office work [14]. A collective of 33 subjects were involved while a set of physiological signals was collected. In [14], Self-Organized Map (SOM) and XY-fused Kohonen Network were used and a classification accuracy of 73.75% achieved for discriminating stress from cognitive load. Another research by [15] used sitting postures to identify emotion expressions. The sitting postures have the semantic factors: "arousal", "pleasantness", and "dominance", so emotion expressions of the sitting postures are like those of the facial expressions [15].

Our proposed approach differs from the existing literature in two main aspects: (1) most of the above works focused their study on office workers but our study, in addition to office workers, targets older people who spend much of their time at home; (2) we employ different algorithms for modeling our posture classifier using several pressure sensors both at the seat and backrest of the chair (32 x 32). Furthermore, in addition to the traditional techniques, we also implemented Deep Neural Network (DNN) algorithm for building predictive models. In the current computerized world, even while involved in leisure or work activity, people tend to spend most of their daily life sitting at computer desks. Therefore, a means to remind people about their sitting habits and provide recommendations that can counterbalance, such as physical exercise, is important. In this work, we propose to identify six sitting postures, which are back, empty, left, right, front and still. An identified sitting posture could potentially be used by an end-user or researcher towards putting solutions to bad sitting posture habits.

## 3 Proposed Method

### 3.1 Data Collection

In this study, pressure sensors were used for collecting the pressure distribution data from a sensor plated chair. The pressure mat was built by members of the Micro Electro-Mechanical-Systems (MEMS) unit of Fondazione Bruno Kessler (FBK) research institute. The dataset was collected at the Intelligent Interfaces and Interaction (I3) research unit of FBK. These mats have a 32 by 32 pressure sensors, which are placed at the seat (bottom) and backrest of the chair. These 64 sensors cover the seat and backrest body of the chair and are able to detect every pressure which is placed on it. In this study, we have tried to investigate whether it is possible to predict sitting postures using the collected information from the pressure sensors. In order to collect both the controlled and realistic datasets, we involved a total of 50 participants, 11 and 39 individuals for each respectively. In the controlled dataset collection, participants were told what positions to hold in a controlled lab setting. Similarly, the realistic dataset were collected in a controlled lab environment by employing a wizard Oz method where participants perform activities to change a television channel by simply moving/changing their body position (moving right corresponds to channel+, moving left for channel-, moving forward for channel page up and moving backward for channel page down). The channels of the television were changed when the button (corresponding to the body movement) was clicked.

### 3.2 Dataset Description

A total of 1980 and 4875 data examples were collected, from two set of participants, for the controlled and realistic datasets, respectively. The controlled dataset was collected from 11 individuals who were told to hold a certain posture for 15 seconds. For every sitting posture, including empty and still positions, data were recorded for the 32 sensor values in every half a second. So, we have 30 snapshots of sensor values. For the realistic dataset collection, we only used five sitting postures excluding the empty posture class. From a total of 39 participants, 25 examples were recorded for each of the five postures resulting in a total of 125 data examples in each participant's file. The dataset was collected within an interval of 0.75 seconds as at time t01, t02, t03, t04 and t05. The assumption in this data collection was that at t01, which is the 1st recurrent element, a participant might not have moved yet or might just be starting to move. At time t05 participant might already be moving back to "center". We found t03, the 3rd recurrent element, and element t234, which is a combination of 2nd, 3rd, and 4th recurrent elements, to be less noisy than the other timestamps, good representatives of the final posture and hence easier to predict and useful for learning.

We used the still sitting posture to normalize both datasets by subtracting it from the other sitting postures. The empty sitting posture data was collected from the empty chair (while there was no one sitting on it). Although the data collected from empty chair were supposed to be empty, there were some sensor reading values. These data were collected from a sensorized chair which was built by members of the Micro Electromechanical-Systems (MEMS) unit of FBK research institute. The chair has a 32 by 32 pressure sensors, which are placed at the seat (bottom) and backrest of the chair as indicated in Fig. 1. These 64 sensors cover the whole body of the chair and able to detect every pressure which is placed on it.

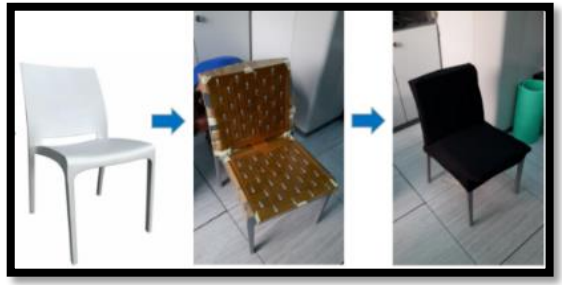

**Fig. 1.** Sensor mounted smart chair.

### 3.3 Data Preprocessing

Since the goal of this study is to investigate whether it is possible to predict sitting postures using data collected from sensor slipped chair (from both the seat and back of the chair), unnecessary columns and invalid sensor readings and outliers are removed from the datasets. The 8x8 matrix where the data collected from each of the 32 sensors is mapped on (see Table 1), does not contain missing values.

Inside each individual file, there are 125 examples collected for each of the five sitting postures. The raw data has a total of 80 features. Since, our goal in this study is to investigate whether it is possible to predict or classify person's sitting posture, we only consider the 64 back and seat sensor features and the class label posture feature. As a result we removed a total of 15 features from the dataset. After we extracted all of the initial features, we then continued to read the sensor values from each individual for each posture and put them into a nested list which we then mapped to an 8x8 matrix for ease of manipulation. Then we merged all individual files into a single Comma-Separated Values file.

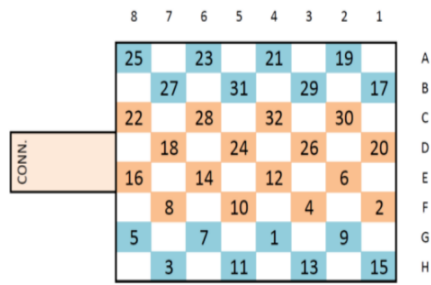

**Fig. 2.** The distribution of the 32 sensors on the printed circuit board

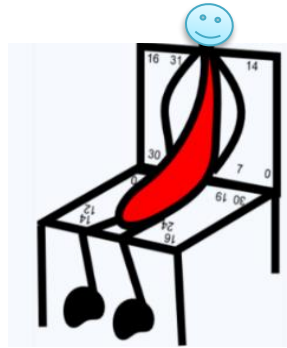

**Fig. 3.** A person seated on a sensor plated smart chair [mirror view]

In the raw collected sensor data, we saw that some of the sensor readings were outliers that deviate from most of the values of a particular sensor in each person's file. In order to handle this, we first calculate the average still value of each sensor column by summing the sensor values in that column for a single person file (which contains 125 data records) and we set a certain threshold. Then we replace the sensor's values which are above the threshold, with the average still value. We did this for the entire 64 seat and back sensors. In addition to this, to avoid discrepancies of sensor readings due to weight variation and fluctuating seating habits of individuals, we normalize each column sensor's for each person with the average still value of the column sensor's that we calculated before. Using these average values, we then subtract it from each of the other posture values of the individual. Normalizing using these still posture values helps to avoid some defects that come due to weight variation between individuals, fluctuating sitting habits and others. The numbers in Table 1 are the positions of the sensors on the printed circuit board of the Texas instrument board. As can be seen in Fig. 2 and Fig. 3, we can visualize the back sensors projection as the face of the sitting person coming out of the mirror.

### 3.4 Feature Generation

In addition to the 64 sensor features, we generate other features that will better describe our dataset and contribute to improve classification accuracy. Since the pressure distribution of sensors on the chair has a great impact, we generate four other features by calculating the center of mass of the 32 sensors of the seat and 32 sensors of the back. We calculated the center of mass for both the seat and back sensors separately. We also extracted features for each of the seat and back sensors dividing the given 32 sensors into four quadrants and edges. Each category in both the quadrant and the edge contains 8 sensor readings. Concerning the quadrant features, each of them are extracted by taking half of the rows and half of the columns from the 8x8 matrix as indicated in Table 2.

**Table 1.** Projection of the seat and back sensors on an 8x8 matrix

|  | Left leg/ right shoulder | | | | Right leg/ left shoulder | | |
|---|---|---|---|---|---|---|---|
|  | Col 1 | Col 2 | Col 3 | Col 4 | Col 5 | Col 6 | Col 7 | Col 8 |
| Row 1 | 16 |  | 31 |  | 9 |  | 14 |  |
| Row 2 |  | 24 |  | 23 |  | 1 |  | 12 |
| Row 3 | 18 |  | 29 |  | 11 |  | 10 |  |
| Row 4 |  | 26 |  | 21 |  | 3 |  | 8 |
| Row 5 | 20 |  | 27 |  | 15 |  | 6 |  |
| Row 6 |  | 28 |  | 17 |  | 5 |  | 4 |
| Row 7 | 22 |  | 25 |  | 13 |  | 2 |  |
| Row 8 |  | 30 |  | 19 |  | 7 |  | 0 |
|  |  |  |  | Bottom |  |  |  |  |

We used the Random Forest (RF) Classifier algorithm to evaluate the importance of each of the features in both the controlled and realistic datasets.

### 3.5 Model Development and Testing

The study is conducted using both the controlled and realistic datasets separately. In both cases, we have conducted various experiments using different combinations such as using normalized and non-normalized dataset, using full sensor features, using selected features, and using selected features with a portion of class labels (two, four, five and all of the classes) with all the five machine learning algorithms used in the study. Specifically, the different experiment settings conducted on the controlled dataset are the following: non-normalized and normalized dataset and 3 other features, non-normalized and normalized sensor and top 10 features, normalized dataset for the whole 55 feature sets, top selected features for non-normalized and normalized datasets, selected features with selected class labels for non-normalized and normalized dataset, and using a DNN on top selected features with the full class labels normalized dataset, five class label normalized dataset with top selected features, and four class label normalized dataset with top selected features.

**Table 2.** Features from the 32 sensors of seat/back divided into four quadrants.

| | Col 1 | Col 2 | Col 3 | Col 4 | Col 5 | Col 6 | Col 7 | Col 8 |
|---|---|---|---|---|---|---|---|---|
| Row 1 | 16 | | 31 | | 9 | | 14 | |
| Row 2 | | 24 | | 23 | | 1 | | 12 |
| Row 3 | 18 | | 29 | | 11 | | 10 | |
| Row 4 | | 26 | | 21 | | 3 | | 8 |
| Row 5 | 20 | | 27 | | 15 | | 6 | |
| Row 6 | | 28 | | 17 | | 5 | | 4 |
| Row 7 | 22 | | 25 | | 13 | | 2 | |
| Row 8 | | 30 | | 19 | | 7 | | 0 |

Like the experiments on the controlled dataset, the following experiment settings were used on the realistic dataset: non-normalized and normalized dataset, experimentation using seat, back or both seat and back sensors dataset, using senior, young or both age groups, using full, 234th and 3rd recurrent elements, experimentation using various length class labels of the posture feature, and using third recurrent element with varying number of classes. All of these experiments are conducted considering number of participants, age group, class label, recurrent element, printed circuit board (seat or backrest), extracted features, and normalized or non-normalized dataset. Unlike the controlled dataset collection (which uses only the 32 sensors of the seat), we use both the 32 by 32 sensors for the realistic dataset collection. In the realistic dataset, five different postures (two before the click and 3 after the click) were collected within five different time frames for each particular posture. Click is a wizard of Oz technique that we employ to perform the position change. As a result, we conduct experiments by using different levels of recurrent elements of the data.

We used five different algorithms and several classifier models are built using RF, Gaussian Naïve Bayes (GNB), Logistic Regression, SVM and DNN. All the models are evaluated using the KFold cross validation technique. Scikit-learn and Tensor-Flow libraries were used. Accuracy of the five classifiers, which were trained with top selected feature sets on the normalized controlled dataset, is reported in Table 3. We train our models with a total of 1800 data examples where 1620 are used for training and the rest 180 are used for testing. As can be seen in table 3, DNN scores superior performance measure compared to all the other methods, be it on the full class labels or a portion of them. The front, left and right postures are relatively easy to predict in all of the classifiers compared to classifying the back posture. Table 4 shows the result of the RF algorithm using the realistic dataset 3$^{rd}$ recurrent element with the left and right class labels.

**Table 3.** Summary classification accuracy results using the normalized controlled dataset

| Selected feature sets (Normalized dataset) | Classifiers | | | | |
|---|---|---|---|---|---|
| | RF | GNB | SVM | LR | DNN |
| Full class label | 82% | 88% | 75% | 81% | 93% |
| Five class label | 86% | 95% | 82% | 88% | 95% |
| Four class label | 85% | 88% | 83% | 89% | 98% |

**Table 4.** Classification report of the RF algorithm using the 3rd recurrent element with left and right class labels normalized realistic dataset.

| | Precision | Recall | F1-score | Support |
|---|---|---|---|---|
| Left | 0.98 | 0.95 | 0.97 | 195 |
| Right | 0.95 | 0.98 | 0.97 | 195 |
| Avg/ total | 0.97 | 0.97 | 0.97 | 390 |

## 4    Results

We have conducted a total of eight experiments in the study using the controlled dataset while we performed a total of six experiments in the study with the realistic dataset using the five classifiers. From all the experiments, the one we did using the DNN algorithm with normalized dataset and four class labels scored the highest result with 98% prediction accuracy. Among the five classifiers, DNN, GNB and RF scored the highest performance in most of our experiments. Although most of the models trained on the controlled dataset were able to classify most of the sitting postures, it was also difficult to easily predict the back posture. The main reason for this might be the usage of only the seat sensors in the data collection, which resulted in the incomplete representation of back postures.

The study using the realistic dataset the highest score was recorded by the RF algorithm with an accuracy score of 97% from a model developed using the third recurrent element and only the left and right class labels as indicated in Table 5. Classifying sitting postures using the full timeframe dataset with all the five class labels was very difficult. As we have seen in our experimentations, the models performance improved when we used varying length of class labels and different portions of the timeframe dataset. The reason for not correctly classifying the full postures might be due to the nature of the dataset as well as the participant's motionless character. Fig. 4 shows the plot of the postures using the 3$^{rd}$ recurrent element with left and right class labels normalized dataset.

**Table 5.** Summary accuracy of the models using the 3rd recurrent element with full/4/3/2 class labels (realistic dataset)

| Recurrent element | | Classifiers | | | | |
|---|---|---|---|---|---|---|
| | | RF | GNB | LR | SVM | DNN |
| 3rd | Full | 77% | 66% | 67% | 69% | 56% |
| | 4 | 77% | 66% | 70% | 71% | 64% |
| | 3 | 87% | 78% | 82% | 84% | 79% |
| | 2 | 97% | 95% | 93% | 94% | 96% |

As it can be observed from the results of the two studies, the one with the controlled dataset seems to perform better than the one with the realistic dataset in classifying sitting postures. Specially, the DNN performed much better in most of the experiments involving the controlled dataset. In addition to this, it was difficult to predict back sitting posture in the case of controlled dataset, but it was relatively easier using the realistic dataset. The reason for this is due to the addition of the 32 back sensors dataset. Variations in the number of class labels does not seem to have that much impact in the case of classifying the controlled dataset while it is important in the realistic dataset.

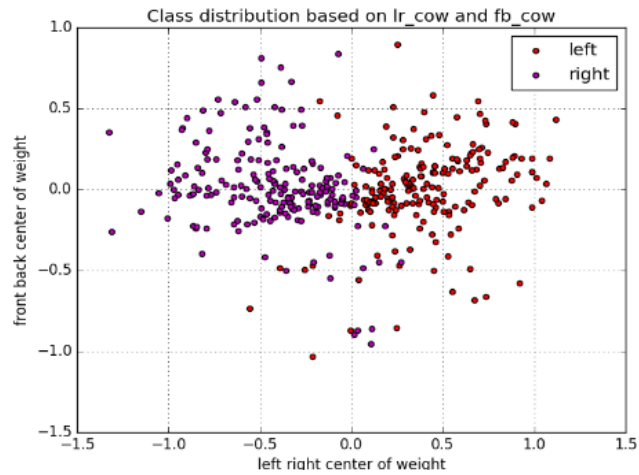

**Fig. 4.** Plot of the postures using only the third recurrent element with left and right class labels normalized realistic dataset.

These differences exist due to the different nature of the two datasets. There is a visible difference in each sensor readings in each of the different data examples in the controlled dataset. This might be associated with a participant performing movements to a certain degree. In the case of the realistic dataset, sensor readings in the different data examples seem closer, meaning participants were in relatively steady sitting conditions compared to the controlled dataset. In addition to the above reasons, we associate performance differences with the variations in dataset size. In the controlled dataset, we have used a total of 1800 data examples while we have a total of 4875 records in the realistic dataset for training and testing the models.

## 5       Conclusion and Recommendations

The main goal of this study was to build a predictive model which can classify the sitting posture of a person and enable individuals to monitor their sitting habits for healthy living. For this purpose, we have collected two datasets from 32 and 64 sensor coated chairs and carried out two separate studies. In addition to the sensor data, we have also generated new features in both studies depending on the center of weight of the seat and back sensors, partitioning the seat and back of the chair into four quadrants and edges.

We have conducted experiments using the sensor and the generated features to build predictive models for both studies. As a result, the model that we built using controlled dataset was able to classify six sitting postures (back, empty, left, right, front and still) with an accuracy score of 93% using a DNN. The highest accuracy score of 98% was achieved in classifying four of the basic sitting postures (back, left, right and front) using DNN. In the second study, our predictive model was able to classify five sitting postures (back, front, left, right and still) with a classification accuracy of 77% using the 3rd recurrent element. The model built on the 3rd recurrent element with only two class labels was able to classify the left and right sitting postures with the highest accuracy score of 97% using RF classifier.

In general, we have achieved good results for both studies in classifying sitting postures, particularly using the controlled dataset. However, performance variation exist between the two studies probably due to differences in dataset size, number of class labels used, data collection method and participants' sitting habit.

In this study, we used a dataset collected within a controlled lab environment, employed limited numbers of participants, and identified six sitting postures. For future work, our method can be extended by collecting real world dataset, employing increased number of participants, and sitting postures. Different sitting activities can also be considered towards developing a full system that can work with variety of chair types and environments.

**Acknowledgements.** This publication has emanated from research supported in part by a grant from Science Foundation Ireland under Grant number 18/CRT/6183. For the purpose of Open Access, the author has applied a CC BY public copyright licence to any Author Accepted Manuscript version arising from this submission'.

# References

1. Roland Zemp, Matteo Tanadini, Stefan Plüss, Karin Schnüriger, Navrag B. Singh, William R. Taylor, Silvio Lorenzetti: Application of Machine Learning Approaches for Classifying Sitting Posture Based on Force and Acceleration Sensors. Hindawi Publishing Corporation BioMed Research International, Zurich, Switzerland (2016).
2. Halender, M.G. and Zhang L.: Field studies of comfort and discomfort in sitting. Ergonomics (1997).
3. Yong Min Kim, Youngdoo Son, Wonjoon Kim, Byungki Jin, Myung Hwan Yun: Classification of Children's Sitting Postures Using Machine Learning Algorithms. Applied Sciences, MDPI, Basel, Switzerland (2018).
4. Hamilton, M.T., Healy, G.N., Dunstan, D.W., Zderic, T.W. and Owen, N.: Too little exercise and too much sitting- inactivity physiology and the need for new recommendations on sedentary behavior. Current Cardiovascular Risk Reports (2008).
5. Owen, N., Healy, G.N., Matthews, C.E. and Dunstan, D.W.: Too much sitting: the population-health science of sedentary behavior. Exercise and sport sciences reviews (2010).
6. Slavomir Matuska, Martin Paralic, Robert Hudec: A Smart System for Sitting Posture Detection Based on Force Sensors and Mobile Application. Hindawi Mobile Information Systems (2020).
7. Wahlström, J.: Ergonomics, musculoskeletal disorders and computer work. Occupational Medicine (2005).
8. Mutlu Bilge, Andreas Krause, Jodi Forlizzi, Carlos Guestrin, Jessica Hodgins: Robust, Low-Cost, Non-Intrusive Sensing and Recognition of Seated Postures. In: Proceedings of the 20th annual ACM symposium on User interface software and technology, pp. 149–158. Carnegie Mellon University, 5000 Forbes Avenue Pittsburgh, PA 15213 USA (2007).
9. Kazuhiro Kamiya, Mineichi Kudo, Hidetoshi Nonaka and Jun Toyama: Sitting posture analysis by pressure sensors. IEEE (2008)
10. Jianquan Wang, Basim Hafidh, Haiwei Dong, Abdulmotaleb El Saddik: Sitting Posture Recognition Using a Spiking Neural Network. IEEE Sensors Journal, University of Rochester (2020).
11. Yong -Ren Huang, Xu-Feng Ouyang: Sitting Posture Detection and Recognition Using Force Sensor. In: 5th International Conference on Biomedical Engineering and Informatics, IEEE (2012).
12. Haeyoon Cho, Hee-Joe Choi, Chae-Eun Lee, Choo-Won Sir: Sitting Posture Prediction and Correction System using Arduino-Based Chair and Deep Learning Model. In: 12th Conference on Service-Oriented Computing and Applications (SOCA), IEEE (2019).
13. Erin Griffiths, T. Scott Saponas, A.J. Bernheim Brush: Health Chair- Implicitly Sensing Heart and Respiratory Rate. UBICOMP'14, Seattle. USA, (2014).
14. Bert Arnrich, Cornelia Setz, Roberto La Marca, Gerhard Troster, Ulrike Ehlert: What does Your Chair Know about Your Stress Level? IEEE Transactions on Information Technology in Biomedicine, vol. 14, (2009).
15. Tatsuya Shibata and Yohei Kijima: Emotion Recognition Modeling of Sitting Postures by using Pressure Sensors and Accelerometers. In: 21st International Conference on Pattern Recognition (ICPR), Tsukuba, Japan (2012).